\def\year{2021}\relax
\documentclass[letterpaper]{article} 
\usepackage{aaai21}  
\usepackage{times}  
\usepackage{helvet} 
\usepackage{courier}  
\usepackage[hyphens]{url}  
\usepackage{graphicx} 
\usepackage{multirow} 
\usepackage[switch]{lineno} 
\usepackage{xcolor} 
\usepackage{placeins} 
\usepackage{natbib}  
\usepackage{caption} 
\title{Many Hands Make Light Work: Using Essay Traits to Automatically Score Essays}
\author {
        Rahul Kumar\textsuperscript{\rm 1},
        Sandeep Mathias\textsuperscript{\rm 2},
        Sriparna Saha\textsuperscript{\rm 3},
        Pushpak Bhattacharyya\textsuperscript{\rm 2}\\
}
\affiliations {
    \textsuperscript{\rm 1} Department of Electrical Engineering, IIT Patna \\
    \textsuperscript{\rm 2} Department of Computer Science and Engineering, IIT Bombay \\
    \textsuperscript{\rm 3} Department of Computer Science and Engineering, IIT Patna \\
    {\tt \{rahul.ee16, sriparna\}@iitp.ac.in \& \{sam, pb\} @cse.iitb.ac.in }
}
\begin{document}
\maketitle

\begin{abstract}
Most research in the area of automatic essay grading (AEG) is geared towards scoring the essay {\em holistically} while there has also been some work done on scoring individual essay traits. In this paper, we describe a way to score essays holistically using a multi-task learning (MTL) approach, where scoring the essay holistically is the primary task, and scoring the essay traits is the auxiliary task. We compare our results with a single-task learning (STL) approach, using both LSTMs and BiLSTMs. We also compare our results of the auxiliary task with such tasks done in other AEG systems. To find out which traits work best for different types of essays, we conduct ablation tests for each of the essay traits. We also report the runtime and number of training parameters for each system. We find that MTL-based BiLSTM system gives the best results for scoring the essay holistically, as well as performing well on scoring the essay traits. 
The MTL systems also give a speed-up of between \textbf{2.30} to \textbf{3.70} times the speed of the STL system, when it comes to scoring the essay and all the traits.
\end{abstract}

\section{Introduction}
\label{Introduction Section}

An \textbf{essay} is a piece of text that is written in response to a topic, called a prompt \cite{mathias-bhattacharyya-2020-neural}. Qualitative evaluation of the essay consumes a lot of time and resources. Hence, in \citeyear{page1966imminence}, \citeauthor{page1966imminence} proposed a method of automatically scoring essays using computers \cite{page1966imminence}, giving rise to the domain of Automatic Essay Grading.

Essay traits are different aspects of the essay that can aid in explaining the score assigned to the essay. Examples of essay traits include content (how much information is present in the essay) \cite{page1966imminence}, organization (how well the essay is structured) \cite{persing-etal-2010-modeling}, style (how well written the essay is) \cite{page1966imminence}, prompt adherence (how much the essay stays on topic for the essay prompt) \cite{persing-ng-2014-modeling}, etc.

Most of the research work done in the field of AEG is geared toward scoring the essay holistically, rather than studying the importance of essay traits in the overall essay score. In this paper, we ask the question:
\begin{center}
\textbf{``Can we use information learnt from scoring essay traits to score an essay holistically?''}
\end{center}

In our paper, not only do we score essays holistically, but we also describe how to score essay traits simultaneously in a multi-task learning framework. Scoring essay traits is essential as it could help in explaining why the essay was scored the way it was, as well as providing valuable insight to the writer about what aspects of the essay were well-written and what the writer needs to improve.

Multi-task learning is a machine learning technique where we use information from multiple auxiliary tasks to perform a primary task \cite{caruana1997multitask}. In our experiments, scoring the individual essay traits is the auxiliary task, and scoring the essay holistically is the primary task.

\paragraph{Contributions.} In this paper, we describe a way to simultaneously score essay traits and the essay itself using multi-task learning. We evaluate our system against different types of essays and essay traits. We also share our code and the data for reproducibility and further research.

\paragraph{Organization of the Paper.} The rest of the paper is organized as follows. The motivation for our work is described in Section \ref{Motivation Section}. We describe related work in Section \ref{Related Work Section}. We describe our system's architecture and dataset in Sections \ref{System Architecture Section} and \ref{Dataset Section} respectively. We describe our experiment setup in Section \ref{Experiment Setup Section}. We report our results and analyze them in Section \ref{Results Section}. Finally, we conclude our paper and describe future work in Section \ref{Conclusion Section}.


\section{Motivation}
\label{Motivation Section}

Most of the work done in the area of automatic essay grading is in the area of holistic AEG - where we provide a single score for the entire essay based on its quality. However, for \textbf{\textit{writers}} of an essay, a holistic score alone would not be enough. Providing trait-specific scores will tell the writer which aspects of the essay need improvement.

In our dataset, we observe that writers of good essays usually have a lot of content, appropriate word choice, very few errors, etc. Essays that are poorly written often lack one or more of these qualities (i.e. they are either too short, have lots of errors, etc.). We, therefore, observe a high correlation between individual trait scores and the overall essay score (Pearson correlation trait scores and overall essay score $>0.7$ across all essay sets in our dataset). Hence, we believe that using essay trait scores will benefit in scoring the essay holistically, as their scores will provide more relevant information to the AEG system.

\section{Related Work}
\label{Related Work Section}

In this section, we describe related work in the area of automatic essay grading and multi-task learning.

\subsection{Holistic Essay Grading}
Holistic essay grading involves assigning an overall score for an essay \cite{mathias-bhattacharyya-2020-neural}. The first AEG system was designed by \citet{page1966imminence}. In the decade of the 2000s there were a lot of AEG systems which were developed commercially (see \citet{shermis2013handbook} for more details).

After the release of Kaggle's Automatic Student Assessment Prize's (ASAP) Automatic Essay Grading (AEG) dataset in 2012\footnote{https://www.kaggle.com/c/asap-aes}, 
there has been a lot of research on holistic essay grading. Initial approaches, such as those of \citet{phandi-etal-2015-flexible} and \citet{zesch-etal-2015-task} used machine learning techniques in scoring the essays. More recent papers look at using a number of deep learning approaches, such as LSTMs \cite{taghipour-ng-2016-neural,tay-2018-skipflow} and CNNs \cite{dong-zhang-2016-automatic} or both \cite{dong-etal-2017-attention,zhang-litman-2018-co,zhang-litman-2020-automated}. \citet{zhang-litman-2020-automated} describe a way to extract important information, called topical components, from a source-dependent response\footnote{We define what a source-dependent response is in the Dataset Section (i.e. Section \ref{Dataset Section}).}.

\subsection{Trait-specific Essay Grading}

In the last decade or so, there has been some work done in scoring essay traits such as sentence fluency \cite{chae-nenkova-2009-predicting},  organization \cite{persing-etal-2010-modeling,taghipour2017robust,mathias-etal-2018-eyes,song-etal-2020-hierarchical}, thesis clarity \cite{persing-ng-2013-modeling,ke-etal-2019-give} coherence \cite{somasundaran-etal-2014-lexical,mathias-etal-2018-eyes}, prompt adherence \cite{persing-ng-2014-modeling}, argument strength \cite{persing-ng-2015-modeling,taghipour2017robust}, stance \cite{persing-ng-2016-modeling}, style \cite{mathias-bhattacharyya-2018-thank} and narrative quality \cite{somasundaran-etal-2018-towards}.
None of the above work, however, uses trait information to score the essay holistically.

There has also been work on scoring multiple essay traits \cite{taghipour2017robust,mathias-bhattacharyya-2018-asap,mathias-bhattacharyya-2020-neural}. \citet{mathias-bhattacharyya-2020-neural} describes work on the use of neural networks for scoring essay traits. Our work combines the scores of essay traits for holistic essay grading.
We focus on using trait-specific essay grading to improve the performance of an automatic essay grading system. We also show how using multi-task learning- simultaneously scoring both the essay and its traits- we are able to speed up the training of our system without too much of a loss in scoring the essay traits.

\subsection{Multi-task Learning}

Multitask Learning was proposed by \citet{caruana1997multitask} where the argument was that training signals from related tasks could help in a better generalization of the model. \citet{collobert2011natural} successfully demonstrated how tasks like Part-of-Speech tagging, chunking and Named Entity Recognition can help each other when trained jointly using deep neural networks. \citet{song-etal-2020-hierarchical} described a multi-task learning approach to score organization in essays, where the auxiliary tasks were classifying the sentences and paragraphs, and the primary task was scoring the essay's organization. \citet{cao-etal-2020-domain} also use a domain adaptive MTL approach to grade essays, where their auxiliary tasks are sentence reordering, noise identification, as well as domain adversarial training. However, they also use all the other essay sets as part of their training, whereas we use only the essays present in the respective essay set for training. 

\begin{figure*}[t]
\centering
\resizebox{\textwidth}{!}{
\includegraphics{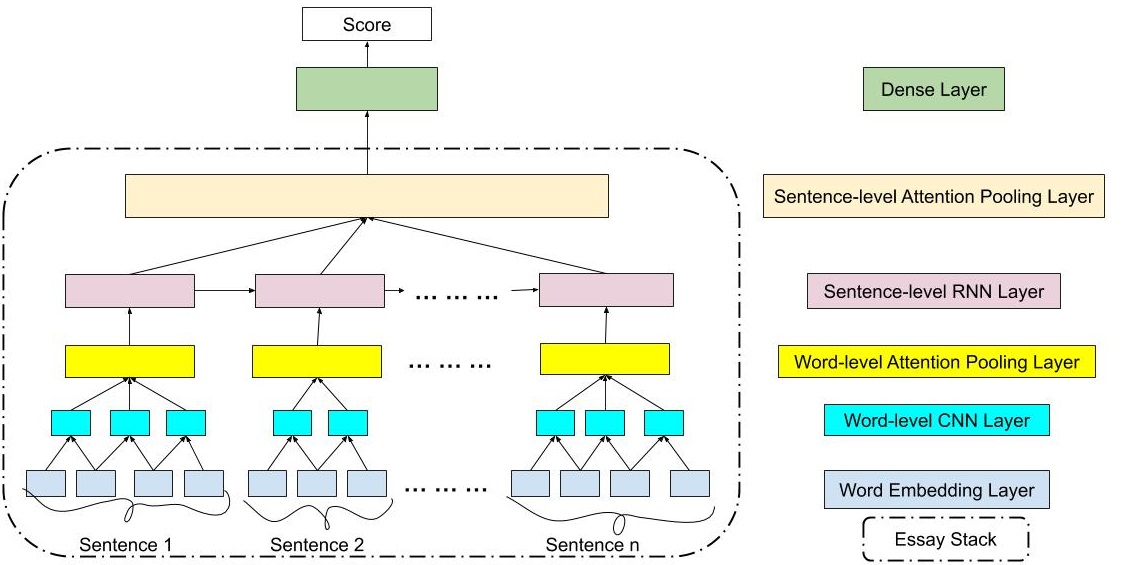}
}
\caption{Essay stack architecture. This is the architecture for the Single-Task Learning systems.}
\label{Essay Stack Architecture}
\end{figure*}

\section{System Architecture}
\label{System Architecture Section}

In this section, we describe the architecture of our system.

\subsection{STL Essay Grading Stack}

\begin{figure*}[t]
\centering
\resizebox{\textwidth}{!}{
\includegraphics{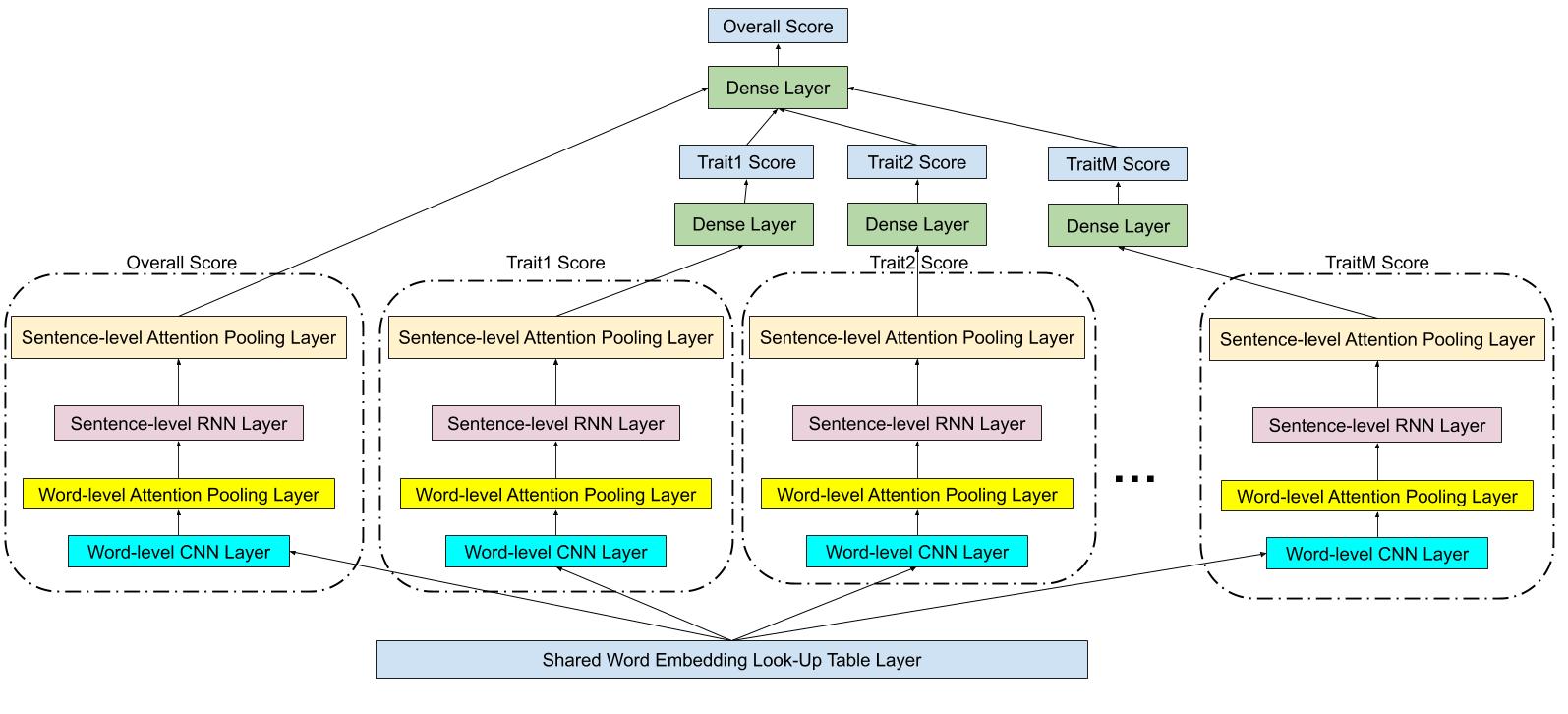}
}
\caption{Architecture of our MTL system showing an input essay with $M$ traits being scored, with the overall score and each trait's essay grading stack.}
\label{MTL Architecture}
\end{figure*}

For scoring the essays, we use essay grading stacks. Each stack is used for scoring a single essay trait. The architecture of the stack is based on the architecture of the holistic essay grading system proposed by \citet{dong-etal-2017-attention}. The essay grading stack takes the essay as input (split into tokens and sentences) and returns the score of the essay / essay trait as the output. Figure \ref{Essay Stack Architecture} shows the architecture for the essay grading stack.

For each essay, we first split the essay into tokens and sentences. This is given as an input to the essay grading stack. In the word embedding layer, we look up the word embeddings of each token. Just like \citet{taghipour-ng-2016-neural}, \citet{dong-etal-2017-attention}, \citet{tay-2018-skipflow}, and \citet{mathias-bhattacharyya-2020-neural}, we use the most frequent 4000 words as the vocabulary with all other words mapping to a special unknown token. This sequence of word embeddings is then sent to the next layer - the 1 dimension CNN layer - to get local information from nearby words. The output of the CNN layer is aggregated using attention pooling to get the sentence representation of the sentence. This is done for every sentence in the essay.

Each of the sentence representations are then sent through a recurrent layer. We experiment on two different types of recurrent layers - a unidirectional LSTM \cite{hochreiter1997long} and bidirectional LSTM (BiLSTM) - as the type of recurrent layer. The outputs of the recurrent layer are pooled using attention pooling to get the representation for the essay. This essay representation is then sent through a fully-connected Dense layer with a sigmoid activation function to score the essay either holistically or a particular essay trait. For our experiments, we minimize the \textbf{mean squared error loss}.

Prior to input, we scale the scores to the range of $[0,1]$ using min-max normalization. The output of the sigmoid function is a scalar in the range of $[0,1]$ which is rescaled back up to a score in the original score range and rounded off to get the score for the essay. This essay stack is used for the scoring of the single-task learning (STL) models.

\subsection{MTL Model}

The architecture of our MTL model for an essay of $M$ traits is shown in Figure \ref{MTL Architecture}. Here, the word embedding layer is \textbf{shared across all the tasks}.
In the multi-task learning framework, each stack is used to learn an essay representation for each essay trait. In a similar manner, the essay representation for the overall score is learnt and it is concatenated with the predicted trait scores before being sent to a Dense layer with a sigmoid activation function to score the essay holistically. For calculating each score - both overall and trait scores - we use the \textbf{mean squared error loss} function. We experimented with multiple weights for the loss function for the essay trait scoring task, but settled on uniform weights for all the traits and the overall scoring task\footnote{This is done because we want to get accurate predictions of the traits scores which are used for predicting the overall score.}.

\section{Dataset Used}
\label{Dataset Section}

\begin{table*}[t]
\centering
\resizebox{\textwidth}{!}{%
\begin{tabular}{|l|c|c|c|c|c|c|}
\hline
\textbf{Essay Set} & \textbf{Score Range} & \textbf{Trait Sc. Range} & \textbf{Word Count} & \textbf{No. of Traits} & \textbf{No. of Essays} & \textbf{Essay Type} \\ \hline
Prompt 1 & 2-12 & 1-6 & 350 & 5 & 1783 & Argumentative / Persuasive \\ 
Prompt 2 & 1-6 & 1-6 & 350 & 5 & 1800 & Argumentative / Persuasive \\ 
Prompt 3 & 0-3 & 0-3 & 100 & 4 & 1726 & Source-Dependent Response \\ 
Prompt 4 & 0-3 & 0-3 & 100 & 4 & 1772 & Source-Dependent Response \\
Prompt 5 & 0-4 & 0-4 & 125 & 4 & 1805 & Source-Dependent Response \\
Prompt 6 & 0-4 & 0-4 & 150 & 4 & 1800 & Source-Dependent Response \\ 
Prompt 7 & 0-30 & 0-6 & 300 & 4 & 1569 & Narrative / Descriptive \\ 
Prompt 8 & 0-60 & 0-12 & 600 & 6 & 723 & Narrative / Descriptive \\ \hline
Total & 0-60 & 0-12 & 100-600 & 4-6 & 12978 & - \\ \hline
\end{tabular}%
}
\caption{Properties of the different essay sets in the ASAP AEG dataset we used in our experiments. Average word count numbers are rounded up to the nearest multiple of 25.}
\label{Dataset Table}
\end{table*}

For our experiments, we use the Automated Student's Assessment Prize (ASAP) Automatic Essay Grading (AEG) dataset. The dataset has a total of 8 essay sets - where each essay set has a number of essays written in response to the same essay prompt. In total, there are nearly 13,000 essays in the dataset. Each of the essays was written by high school students belonging to classes 7 to 10.

Table \ref{Dataset Table} gives the properties of each of the essay sets in our dataset. It reports the overall essay scoring range, traits scoring, average word count, number of traits, number of essays and essay type.

We use the overall scores directly from the ASAP AEG dataset. Since the original dataset only provided trait-specific scores for Prompts 7 \& 8, we use the trait-specific scores provided by \citet{mathias-bhattacharyya-2018-asap}.

\begin{table*}[t]
\centering
\begin{tabular}{|l|c|c|c|c|c|c|}
\hline
\textbf{Essay Set} & \textbf{Trait 1} & \textbf{Trait 2} & \textbf{Trait 3} & \textbf{Trait 4} & \textbf{Trait 5} & \textbf{Trait 6} \\ \hline
Prompt 1 & Content & Organization & Word Choice & Sentence Fluency & Conventions & N/A \\
Prompt 2 & Content & Organization & Word Choice & Sentence Fluency & Conventions & N/A\\
Prompt 3 & Content & Prompt Adherence & Language & Narrativity & N/A & N/A \\
Prompt 4 & Content & Prompt Adherence & Language & Narrativity & N/A & N/A \\
Prompt 5 & Content & Prompt Adherence & Language & Narrativity & N/A & N/A \\
Prompt 6 & Content & Prompt Adherence & Language & Narrativity & N/A & N/A \\
Prompt 7 & Content & Organization & Style & Conventions & N/A & N/A \\
Prompt 8 & Content & Organization & Voice & Word Choice & Sentence Fluency & Conventions \\ \hline
\end{tabular}%
\caption{Traits that are present in each essay set in our dataset. The trait scores are taken from the original ASAP dataset, as well as from ASAP++ \citet{mathias-bhattacharyya-2018-asap}.}
\label{Traits Table}
\end{table*}

Depending on the type of prompt for the essay set, each essay set has a different set of traits. Argumentative / Persuasive essays are essays which the writer is prompted to take a stand on a topic and argue for their stance. These essay sets have traits like content, organization, word choice, sentence fluency, and conventions. Source-dependent responses \cite{zhang-litman-2018-co} are essays where the writer reads a piece of text and answers a question based on the text that they just read\footnote{A sample prompt is ``Based on the excerpt, describe the obstacles the builders of the Empire State Building faced in attempting to allow dirigibles to dock there. Support your answer with relevant and specific information from the excerpt.'' It involves the writer reading the excerpt from \textit{The Empire State Building} by Marcia Amidon Lusted before writing the essay.}. These essay sets have traits like content, prompt adherence \cite{persing-ng-2014-modeling}, language and narrativity \cite{somasundaran-etal-2018-towards}. Narrative / Descriptive essays are essays where the writer has to narrate a story or incident or anecdote. They have traits like content, organization, style, conventions, voice, word choice, and sentence fluency\footnote{Neither the original ASAP dataset, nor \citet{mathias-bhattacharyya-2018-asap} have scored narrativity for the narrative essays.}. Table \ref{Traits Table} lists the different essay traits for each essay set.

\section{Experiments}
\label{Experiment Setup Section}
In this section, we describe our methodology and evaluation metric, as well as experiment configurations and network hyper-parameters.

\subsection{Evaluation Metric}

We use Cohen's Kappa with quadratic weights \cite{cohen1968weighted} (QWK) as the evaluation metric. This is done for the following reasons. Firstly, the final scores predicted by the system are distinct numbers/grades, rather than continuous values; so we cannot use the Pearson Correlation Coefficient or Mean Squared Error. Secondly, evaluation metrics like F-Score and accuracy do not take into account chance agreements. For example, if we are to grade every essay with the mean score or most frequent score, we would get F-Score and accuracy as high as 60\% or more, whereas the Kappa score will be 0! Thirdly, the fact that the scores given are \textbf{\textit{ordered}} (i.e. $0 < 1 < 2 < 3 ...$) means that we need to use weighted Kappa to capture the distance between the actual and predicted scores. Between linear weighted Kappa and QWK, we choose QWK because it rewards matches and punishes mismatches more distinctly than linear weighted Kappa.

\begin{table}[t]
\centering
\resizebox{\columnwidth}{!}{
\begin{tabular}{|l|l|c|}
\hline
\textbf{Layer} & \textbf{Param. Name} & \textbf{Param. Value} \\ \hline
\multirow{2}{*}{Embedding} & Embedding Dim. & 50 \\
 & Embeddings & GloVe \\ \hline
\multirow{2}{*}{Word CNN} & Window Size & 5 \\
 & Filters & 100 \\ \hline
Sentence LSTM & Hidden Units & 100 \\ \hline
 & Epochs & 100 \\
 & Batch Size & 100 \\
 & Dropout Rate & 0.5 \\
 & Initial Learning Rate & 0.001 \\
 & Momentum & 0.9 \\
 & Optimizer & RMSProp \\
\hline
\end{tabular}
}
\caption{Neural network hyper-parameters for each layer, showing the hyper-parameter name and its corresponding value.}
\label{Network Hyperparameters Table}
\end{table}

\subsection{Evaluation Method}

We evaluate our experiments using \textbf{five-fold cross validation}, where, for each essay set, we use \textbf{60\%} of the data as training data, \textbf{20\%} as development/validation data, and the remaining \textbf{20\%} as testing data. We use the same data splits as used by \citet{taghipour-ng-2016-neural}. To avoid overfitting, for each fold, we choose the model which gives the best result on the validation set for evaluating on the test set.

\begin{table*}[t]
\centering
\begin{tabular}{|l|ccc|cc|c|}
\hline
\textbf{Essay Set} & \textbf{Kernel} & \textbf{STL-LSTM} & \textbf{STL-BiLSTM} & \textbf{MTL-LSTM} & \textbf{MTL-BiLSTM} & \textbf{\citet{tay-2018-skipflow}} \\ \hline
Prompt 1 & 0.804 & 0.813 & 0.818 & 0.830* & 0.831* & \textbf{0.832} \\
Prompt 2 & 0.687 & 0.660 & 0.658 & 0.667 & \textbf{0.689}*$\star$ & 0.684 \\
Prompt 3 & \textbf{0.704} & 0.661 & 0.653 & 0.644 & 0.687*$\star$ & 0.694 \\
Prompt 4 & 0.743 & 0.790 & 0.780 & 0.786 & \textbf{0.798} & 0.788 \\
Prompt 5 & 0.799 & 0.798 & 0.789 & 0.782 & 0.800 & \textbf{0.815} \\
Prompt 6 & 0.753 & 0.807 & 0.803 & 0.806 & \textbf{0.813}*$\star$ & 0.810 \\
Prompt 7 & 0.698 & 0.792 & 0.786 & 0.791 & 0.795$\star$ & \textbf{0.800} \\
Prompt 8 & 0.552 & 0.678 & 0.697 & 0.679 & \textbf{0.699}*$\star$ & 0.697 \\ \hline
\textbf{Mean QWK} & 0.717 & 0.750 & 0.748 & 0.748 & \textbf{0.764}*$\star$ & \textbf{0.764} \\
\hline
\end{tabular}
\caption{Results of our experiments for scoring the essays holistically. The first column is the different essay sets. The next three columns represent the baseline results for holistic essay grading using single-task learning. Next two columns are results using multi-task learning. The final column is the results as reported from another state-of-the-art system by \citet{tay-2018-skipflow}. Figures in \textbf{boldface} represent the \textbf{best results} per essay set. * represents a statistically significant improvement using the MTL systems over the STL-LSTM system. $\star$ represents a statistically significant improvement of using the MTL-BiLSTM system over the MTL-LSTM system.}
\label{Overall Score Results Table}
\end{table*}

\subsection{Network Hyperparameters}


Table \ref{Network Hyperparameters Table} gives the different hyperparameters used in our systems. For the sake of uniformity, we use these hyperparameters irrespective of the network configuration (STL vs MTL, or LSTM vs BiLSTM).

We use the \textbf{GloVe} \cite{pennington-etal-2014-glove} pre-trained word-embeddings, trained on the Wikpedia 2014 + Gigaword 5 corpus (6 billion tokens, 400K vocabulary, uncased \textbf{50 dimensions})\footnote{We used GloVe instead of BERT \cite{devlin-etal-2019-bert} due to the fact that BERT's training cost was far too large with little or no increase in performance for AEG \cite{mayfield-black-2020-fine}.}. For the Word-level CNN layer, we use a \textbf{window size of 5}, with a \textbf{filter of size 100} and for the Sentence-level recurrent layer (LSTM / BiLSTM), we have \textbf{100 hidden units}. We run our experiments for \textbf{100 epochs} over a \textbf{batch size of 100}.

We use the \textbf{RMSProp Optimizer} \cite{dauphin2015equilibrated} with an \textbf{initial learning rate of 0.001} and a \textbf{momentum of 0.9} and a \textbf{dropout rate of 0.5}.

\subsection{Experimental Configurations}

To evaluate the performance of our systems in scoring the essay overall, we use 4 different configurations- \textbf{STL-LSTM}, \textbf{STL-BiLSTM}, \textbf{MTL-LSTM}, and \textbf{MTL-BiLSTM}. In addition to the above systems, we also compare our approach with a state-of-the-art string kernel system designed by \citet{cozma-etal-2018-automated}, using the same splits for training, testing, and validation\footnote{\citet{cozma-etal-2018-automated} do not provide their folds, so we run their system on our training/validation/test split, as given by \citet{taghipour-ng-2016-neural}.}.

In the STL configurations, we train our system to predict a single score at a time- either the overall essay score or the score for any of the essay traits. In the MTL configurations, our system learns to score the essay and \textbf{all its traits} simultaneously. The LSTM configurations use only a forward direction LSTM, while the BiLSTM configurations use a bidirectional (i.e. forward and reverse) LSTM.

\section{Results and Analysis}
\label{Results Section}

In this section, we report our results and analyze them for different experiments. 

\subsection{Performance on Holistic Essay Scoring}

Table \ref{Overall Score Results Table} gives the QWK scores of each of our systems as they score each essay set holistically. The different systems used are the Single Task Learning (STL) (only scoring the essay overall) and Multitask Learning (MTL) (scoring the essay and the traits simultaneously). The first column lists out the different essay sets (Prompts 1 to 8). The next three columns report results for STL using both LSTM and BiLSTM, as well as results using the string kernel-based approach of \citet{cozma-etal-2018-automated}. The next two columns report results for the MTL systems using both LSTM and BiLSTM. The last column shows the results reported in a state-of-the-art system using neural networks by \citet{tay-2018-skipflow}.

From the table, we see that the MTL-BiLSTM performs the best of all the systems (as good as the results of \citet{tay-2018-skipflow}). In order to see if the improvements observed are statistically significant, we run the Paired T-Test for each of the essay sets and compare the results using a p-value of $p<0.05$.

We test each of the essay sets in both the MTL systems - MTL-LSTM and MTL-BiLSTM - against the best-performing STL system which we implemented, the STL-LSTM system\footnote{We do not compare it with \citet{tay-2018-skipflow}'s system since they have not released their code.}. We mark results which are statistically significant with a * next to the QWK value in Table \ref{Overall Score Results Table}. Similarly, we also test whether or not the improvements we achieved using the BiLSTM layer instead of the LSTM layer are statistically significant. We mark improvements which are statistically significant in the MTL-BiLSTM column, over the MTL-LSTM column with a $\star$. 

\subsection{Performance on Scoring Essay Traits}

\begin{figure}[ht]
\centering
\includegraphics[width=\columnwidth]{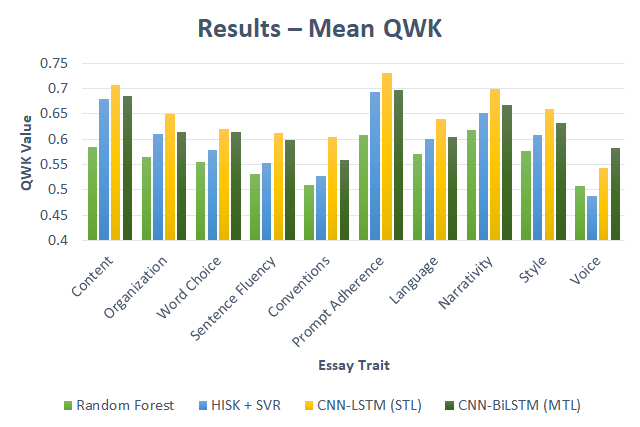}
\caption{Results of the performance of scoring traits for different systems. \textbf{NOTE} that the y-axis starts from a QWK of 0.4. This was done to highlight the difference in the performance of each system for each trait.}
\label{Traits Mean QWK}
\end{figure}

We also look at how our system performs in the auxiliary tasks - namely scoring the different essay traits. Figure \ref{Traits Mean QWK} gives the results of our experiments in scoring the essay traits, using a Random Forest \cite{mathias-bhattacharyya-2018-asap}, the String Kernel (HIST + SVR) \cite{cozma-etal-2018-automated}, CNN-LSTM (STL) \cite{dong-etal-2017-attention}, and Our System (CNN-BiLSTM (MTL)). We use the same evaluation method, which we used for scoring essay traits, with the same data splits. For the STL systems, we train them for every essay trait individually.

We compare the results with that of our MTL-BiLSTM system, which was trained to score the essay traits as auxiliary tasks. Figure \ref{Traits Mean QWK} gives the results of our experiments. From the figure, we see that, while the STL-LSTM system is able to outperform our MTL-BiLSTM system, for most of the traits, the MTL-BiLSTM system performs about 97\% as good as the STL-LSTM system. The main reason for this difference in performance is that the STL-LSTM system tries to optimize for scoring the trait alone, while our MTL system tries to optimize for scoring the essay holistically.

\subsection{Ablation Tests}


\begin{table*}[t]
\centering
\begin{tabular}{|l|c|c|c|c|c|c|c|c|}
\hline
\textbf{Essay Trait} & \textbf{Prompt 1} & \textbf{Prompt 2} & \textbf{Prompt 3} & \textbf{Prompt 4} & \textbf{Prompt 5} & \textbf{Prompt 6} & \textbf{Prompt 7} & \textbf{Prompt 8} \\
\hline
Content & \underline{\textbf{ 0.0148}} & 0.0092 & 0.0064 & 0.0074 & 0.0030 & \underline{\textbf{0.0128}} & \underline{\textbf{0.0102}} & 0.0102 \\
Organization & 0.0122 & 0.0088 & --- & --- & --- & --- & 0.0090 & 0.0052 \\
Word Choice & 0.0080 & \underline{\textbf{0.0164}} & --- & --- & --- & --- & --- & \underline{\textbf{0.0264}} \\
Sentence Fluency & 0.0086 & 0.0036 & --- & --- & --- & --- & --- & 0.0196 \\
Conventions & 0.0090 & 0.0018 & --- & --- & --- & --- & 0.0076 & 0.0056 \\
Prompt Adherence & --- & --- & \underline{\textbf{0.0282}} & \underline{\textbf{0.0112}} & 0.0026 & 0.0044 & --- & --- \\
Language & --- & --- & 0.0080 & 0.0108 & 0.0088 & 0.0030 & --- & --- \\
Narrativity & --- & --- & 0.0092 & 0.0050 & \underline{\textbf{0.0124}} & 0.0062 & --- & --- \\
Style & --- & --- & --- & --- & --- & --- & 0.0030 & --- \\
Voice & --- & --- & --- & --- & --- & --- & --- & 0.0094 \\
\hline
\end{tabular}
\caption{Results of the ablation tests. The numbers show the \textbf{drop in performance} when we ablate each of the essay traits from each of the essay sets (Prompt 1 to 8). The most important features in each essay set are written in \underline{\textbf{boldface and underlined}}. Cells with a --- in them mean that the essay trait was not present in that essay set.}
\label{Ablation Results Table}
\end{table*}

In order to know which trait is most important for each essay set, we run a series of ablation tests. For each essay set, we ablate one essay trait at a time before scoring the essay. Table \ref{Ablation Results Table} reports the results of the ablation test. The values in the table correspond to the \textbf{drop in performance} in scoring the essay holistically. We find that the Content is the most important essay trait for 3 of the essay sets. Prompt Adherence and Word Choice are the most important traits for 2 of the essay sets where they are scored.

\subsection{Error Analysis}

As we have seen, the MTL model generally helps over the STL model when it comes to holistic essay scoring, especially if there is no well-defined rule (Example: Holistic Score = Sum of trait scores) for scoring the essay holistically.

A possible scenario where STL {\em could} help over MTL is if the holistic score is a well-defined function of the trait scores \textbf{AND} the STL system can predict the trait scores with a good deal of accuracy. The essay sets corresponding to Prompts 7 \& 8 are two such essay sets, where the overall score is a function of the individual trait scores.
To verify this, we ran the experiments in a pipelined manner - first scoring the essay traits, then calculating the holistic score using the predicted trait scores and comparing it with the gold standard holistic scores. We found no difference in QWK for Prompt 7 (a QWK of 0.796 vs. 0.795), but a much lesser performance with Prompt 8 (a QWK of 0.684 vs. 0.699) as compared to our MTL-based system. One of the main reasons for this is due to the poor performance in predicting the trait scores as single tasks.

\subsection{Runtime Analysis}


\begin{table}[h]
\centering
\resizebox{\columnwidth}{!}{
\begin{tabular}{|l|c|c|c|}
\hline
\textbf{System} & \textbf{STL Time} & \textbf{MTL Time} & \textbf{Speed-Up} \\
\hline
LSTM & 24.62 hours & 10.45 hours & 2.30 \\
BiLSTM & 40.98 hours & 11.32 hours & 3.70 \\
\hline
\end{tabular}
}
\caption{Total training time for each system for all prompts, traits and folds, using our neural network systems.}
\label{Runtime Table}
\end{table}

\begin{table}[h]
\centering
\begin{tabular}{|l|c|c|}
\hline
\textbf{System} & \textbf{Average} & \textbf{Range} \\ \hline
STL-LSTM & 326K & 326K \\
STL-BiLSTM & 436K & 436K \\
MTL-LSTM & 891K & 829K - 1.08M\\
MTL-BiLSTM & 1.5M & 1.38M - 1.85M\\ \hline
\end{tabular}
\caption{Average and range of training parameters per essay set for each system.}
\label{Training Parameters Table}
\end{table}

We also ran experiments to see how much resources and time our approaches will take. Table \ref{Runtime Table} gives the total training time (in hours). The total training time is the total time taken to train our system to score the essay holistically \textit{as well as all the traits} in that essay set for all 100 epochs. We also report the speed-up when using the MTL approach as compared to the STL approach. From our results, we observe a \textbf{2.30} to \textbf{3.70} speed-up in using the MTL models as compared to using the STL models.

We also report the average number of training parameters per system in Table \ref{Training Parameters Table}. For the STL systems, the number of trainable parameters is the same irrespective of essay set. For the MTL systems, the models, the number of training parameters varies based on the number of essay traits in the essay set. Prompts 3 to 7, which have only 4 traits, have about 1.38 million training parameters. On the other hand, Prompt 8, which has 6 essay traits, has over 1.85 million training parameters.

All our experiments were run on an nVidia GeForce GTX 1080 Ti Graphics Card, using Python version 3.5.2, Keras version 2.2.4 and Tensorflow version 1.14. The essays are tokenized and sentence split using NLTK version 3.4.5.

\section{Conclusion and Future Work}
\label{Conclusion Section}

In this paper, we described an approach to use multi-task learning to automatically score essays and their traits. We achieve this by concatenating a representation of the essay with the trait scores - predicted as an auxiliary task. We compared our results with single-task learning models as well. We found out that the MTL system with the Bi-Directional LSTM outperforms the STL-based systems and has results comparable with the state-of-the-art system of \citet{tay-2018-skipflow}. Next, we evaluated our system's performance in scoring individual essay traits and found that its performance is close to that of the STL systems. We then ran an ablation test and found out which essay trait was important for the corresponding essay sets. We also report our system's performance, which shows a \textbf{2.30} to \textbf{3.70} speed-up of using the multi-task learning system, compared to using a single task learning system.


An exciting avenue of future work is using trait scoring to aid in providing \textit{text feedback} to the writer, like showing where the low score for the trait originates, similar to \citet{hellman-etal-2020-multiple} (for content scoring), rather than a trait-specific score only. We also plan to explore how we can use this approach in the area of \textbf{cross-domain AEG}, where we train our system using essays written in response to one prompt, and test it on essays written in another prompt.
\section*{Ethics Statement}

This submission protects the privacy of the writers of the essays. All named entities, dates, numbers, etc. which could identify the students who wrote the almost 13,000 essays as part of the data set are anonymized (as per the original ASAP AEG dataset). No attempt has been made to solicit their identities.

The trait-specific data, collected by \citet{mathias-bhattacharyya-2018-asap} was also collected in an ethical manner. Annotators for that task were adequately compensated INR 5 per essay for annotation.
\bibliography{aaai21}
\bibliographystyle{aaai2021}
\end{document}